\documentclass[onecolumn,preprintnumbers,amsmath,amssymb,showkeys]{revtex4}
\usepackage{hyperref} 
\usepackage{graphicx}
\pdfoutput=1

\newcommand{\beq}{\begin{eqnarray}}
\newcommand{\eeq}{\end{eqnarray}}
\newcommand{\bfig}{\begin{figure}}
\newcommand{\efig}{\end{figure}}
\newcommand{\bcen}{\begin{center}}
\newcommand{\ecen}{\end{center}}
\newcommand{\incgraph}[1]{\includegraphics[width=\textwidth]{{#1}}}
\newcommand{\fref}[1]{Fig. \ref{#1}}

\newcommand{\pdfrac}[2]{{#1}/\left({#2}\right)}

\newcommand{\bS}{{\bf{S}}}
\newcommand{\pbarj}{\bar{p}_j}
\newcommand{\etaj}{\eta_j}
\newcommand{\epsj}{\varepsilon_j}
\newcommand{\beps}{{\bf{\varepsilon}}}

\newcommand{\half}{\frac{1}{2}}

\newcommand{\si}{\sum_{i=1}^{\infty}}
\newcommand{\sj}{\sum_{j=1}^N}
\newcommand{\pio}{\pi_1}
\newcommand{\pit}{\pi_2}
\newcommand{\po}{p_{1j}}
\newcommand{\bpo}{{\bf{p}}_{1}}
\newcommand{\pt}{p_{2j}}
\newcommand{\bpt}{{\bf{p}}_{2}}
\newcommand{\bp}{{\bf{p}}}

\begin{document}

\title{A non-negative expansion for small Jensen-Shannon Divergences}

\author{Anil Raj}
 \affiliation{Department of Applied Physics and Applied Mathematics\\ Columbia University, New York}
 \email{ar2384@columbia.edu}
\author{Chris H. Wiggins}
 \affiliation{Department of Applied Physics and Applied Mathematics\\ 
		Center for Computational Biology and Bioinformatics\\ Columbia University, New York}
 \email{chris.wiggins@columbia.edu}

\date{\today}

\begin{abstract}
In this report, we derive a non-negative series expansion for the Jensen-Shannon divergence (JSD)
between two probability distributions. This series expansion is shown to be useful for numerical calculations
of the JSD, when the probability distributions are nearly equal, and for which, consequently, small numerical errors dominate evaluation.

\end{abstract}

\keywords{entropy, JS divergence}

\maketitle

\section{Introduction}
The Jensen-Shannon divergence (JSD) has been widely used as a dissimilarity measure between weighted probability
distributions. The direct numerical evaluation of the exact expression for the JSD
(involving difference of logarithms), however, leads to numerical errors when the distributions are 
close to each other (small JSD);
when the element-wise difference between the distributions is $O(10^{-1})$, this naive formula
produces erroneous values (sometimes negative) when used for numerical calculations. In this report, we derive a provably non-negative series expansion for the JSD which can be used in the small JSD limit, where the naive formula fails.

\section{Series expansion for Jensen-Shannon divergence}
Consider two discrete probability distributions $\bpo$ and $\bpt$ over a sample space
$\bS$ of cardinality $N$ with 
relative normalized weights $\pio$ and $\pit$ between them. The JSD
between the distributions is then defined as \cite{lin:1991}
\beq
\Delta_{naive}[\bpo,\bpt;\pio,\pit] = H[\pio\bpo + \pit\bpt] - (\pio H[\bpo] + \pit H[\bpt])
\eeq
where the entropy (measured in nats) of a probability distribution is defined as
\beq
H[\bp] = - \sj h(p_j) = -\sj p_j\log(p_j).
\eeq

Defining 
\begin{eqnarray}
\begin{matrix}
\pbarj & = & ({\po+\pt})/{2}  &;& 0 \leqslant \pbarj \leqslant 1  &;\sj \pbarj = 1 \\
\etaj & = & ({\po-\pt})/{2}   &;& \sj \etaj = 0 &\\
\epsj & = & ({\etaj})/{\pbarj} &;& -1 \leqslant \epsj \leqslant 1 &\\
\alpha & = & \pio-\pit &;& -1 \leqslant \alpha \leqslant 1 &
\end{matrix}
\end{eqnarray}
we have 
\beq
h(\pio\po+\pit\pt) & = & -(\pio(\pbarj+\etaj)+\pit(\pbarj-\etaj)) 
\log (\pio(\pbarj+\etaj)+\pit(\pbarj-\etaj))\notag \\
& = & -\pbarj (1+\alpha\epsj) \left[ \log(\pbarj) + \log(1+\alpha\epsj) \right]
\eeq
and
\beq
\pio h(\po) + \pit h(\pt) & = & -\pio(\pbarj+\etaj) \log (\pbarj+\etaj)
-\pit(\pbarj-\etaj) \log (\pbarj-\etaj)\notag \\
& = & -\half \pbarj(1+\alpha)(1+\epsj) \log (\pbarj(1+\epsj)) 
-\half \pbarj(1-\alpha)(1-\epsj) \log (\pbarj(1-\epsj))\notag \\
& = & -\pbarj (1+\alpha\epsj) \log (\pbarj)
-\half \pbarj(1+\alpha\epsj) \log(1-\epsj^2)
-\half \pbarj(\alpha+\epsj) \log \left(\frac{1+\epsj}{1-\epsj} \right).
\eeq
Thus,
\beq
h(\pio\po+\pit\pt)-(\pio h(\po) + \pit h(\pt)) & = & 
\half \pbarj 
\left[(1+\alpha\epsj) \log \left( \frac{1-\epsj^2}{(1+\alpha\epsj)^2} \right)
+ (\alpha + \epsj) \log \left( \frac{1+\epsj}{1-\epsj} \right) \right].
\eeq

The Taylor series expansion of the logarithm function is given as
\beq
\log (1+x) = \si c_i x^i ; ~~~c_i = \frac{(-1)^{i+1}}{i}.
\eeq
The logarithms in the expression for the J-S divergence can then be written as 
\beq
\log (1+\epsj) & = & \si c_i \epsj^i\notag \\
\log (1-\epsj) & = & \si (-1)^i c_i \epsj^i \\
\log (1+\alpha\epsj) & = & \si c_i \alpha^i \epsj^i.\notag
\eeq

We then have
$\Delta=\half \sj \pbarj \delta_j$, with
\beq \label{eq:expand}
\delta_j&=& (1+\alpha\epsj) \left[ \log (1+\epsj) + \log (1-\epsj) - 2 \log (1+\alpha\epsj) \right]
+ (\alpha+\epsj) \left[ \log (1+\epsj) - \log (1-\epsj) \right]\notag \\
& = & (1+\alpha\epsj) \left[ \si c_i \epsj^i + \si (-1)^i c_i \epsj^i - 2 \si c_i \alpha^i \epsj^i \right]
+ (\alpha+\epsj) \left[ \si c_i \epsj^i - \si (-1)^i c_i \epsj^i \right]\notag \\
& = & \si c_i \left[ \epsj^i + \alpha \epsj^{i+1} + (-1)^i \epsj^i + (-1)^i \alpha \epsj^{i+1} 
-2 \alpha^i \epsj^i -2 \alpha^{i+1} \epsj^{i+1} + \alpha \epsj^i + \epsj^{i+1} 
+ (-1)^{i+1} \alpha \epsj^i + (-1)^{i+1} \epsj^{i+1} \right]\notag \\
& = & \si c_i \left[ \left\{ (-1)^i - 2 \alpha^i + \alpha + (-1)^{i+1} \alpha + 1 \right\} \epsj^i 
+ \left\{ (-1)^i \alpha - 2 \alpha^{i+1} + 1 + (-1)^{i+1} + \alpha \right\} \epsj^{i+1} \right].
\eeq

When $i=1$, ${\rm coeff}(\epsj) = c_1 (-1 -2\alpha + \alpha + \alpha + 1) = 0$. 
The first non-vanishing term
in the expansion is
then of order 2. Shifting indices of the first term in Eqn. \eqref{eq:expand} gives
\beq
\delta_j & = & \si \left[ c_{i+1} \left\{ (-1)^{i+1} - 2 \alpha^{i+1} + \alpha + (-1)^{i+2} \alpha + 1 \right\}
+ c_i \left\{ (-1)^i \alpha - 2 \alpha^{i+1} + 1 + (-1)^{i+1} + \alpha \right\} \right] \epsj^{i+1}\notag \\
& = & \si (c_i + c_{i+1}) \left\{ (-1)^i \alpha - 2 \alpha^{i+1} + \alpha + 1 + (-1)^{i+1}) \right\} \epsj^{i+1}\notag \\
& = & \si \frac{(-1)^{i+1}}{i(i+1)} \left\{ (-1)^i \alpha - 2 \alpha^{i+1} + \alpha + 1 + (-1)^{i+1}) \right\} \epsj^{i+1}\notag \\
& = & \si B_i \epsj^{i+1}
\eeq
where
\beq
B_i & = & \frac{1 - \alpha + (-1)^{i+1} (1+ \alpha - 2\alpha^{i+1})}{i(i+1)}
=\Biggl\{
\begin{matrix}
\pdfrac{2(1-\alpha^{i+1})       }{i(i+1)}&i~{\rm odd},\cr
\pdfrac{-2 (\alpha-\alpha^{i+1})}{i(i+1)}&i~{\rm even.}
\end{matrix}
\eeq

This series expansion can be further 
simplified
as
\beq
\delta_j & = & \si \left( B_{2i-1} + B_{2i} \epsj \right) \epsj^{2i}\notag \\
& = & \si B_{2i-1} \left( 1+\frac{B_{2i}}{B_{2i-1}} \epsj\right) \epsj^{2i}, \\
\frac{B_{2i}}{B_{2i-1}} \epsj & = & - \left( \frac{2i-1}{2i+1} \right) \alpha \epsj.
\eeq
Since $-1 \leqslant \alpha \epsj \leqslant 1$, we have $-1 \leqslant \frac{B_{2i}}{B_{2i-1}} \epsj \leqslant 1$.
Thus, for every $i$, $(B_{2i-1} + B_{2i} \epsj) \epsj^{2i} \geqslant 0$, making $\delta_j$ --- 
and the series expansion for $\Delta_{naive}$ --- non-negative up to all orders.

\section{Numerical Results}
The accuracy of the truncated series expansion can be compared with the naive formula
by measuring the JSD between randomly generated probability distributions.
Pairs of probability distributions with $-4 \leqslant \log_{10} \| \beps \| < 0$, where
$\| \beps \| = \sqrt{\frac{\sj \epsj^2}{N}}$, were randomly generated and the J-S divergence
between each pair was calculated by both a direct evaluation of the exact expression ($\Delta_{naive}$)
and the approximate expansion ($\Delta_k; k \in \{3,6,9,12\}$), where 
\begin{eqnarray}
\Delta_k = \half \sj \pbarj \delta_{jk}\quad ; \quad \delta_{jk} = \sum_{i=1}^k B_i \epsj^{i+1}.
\end{eqnarray} 
The results shown in \fref{compare} suggest the 
series expansion to be a more numerically useful formula when the probability distributions differ 
by $\| \beps \| \sim O(10^{-0.5})$. \fref{negjs} further shows that when $\| \beps \| \sim O(10^{-6})$, 
a direct evaluation of the exact formula for JSD gives negative values (when implemented
in \textsc{matlab}).  

\bfig
\begin{minipage}[t]{0.485\linewidth}
\incgraph{JScompare}
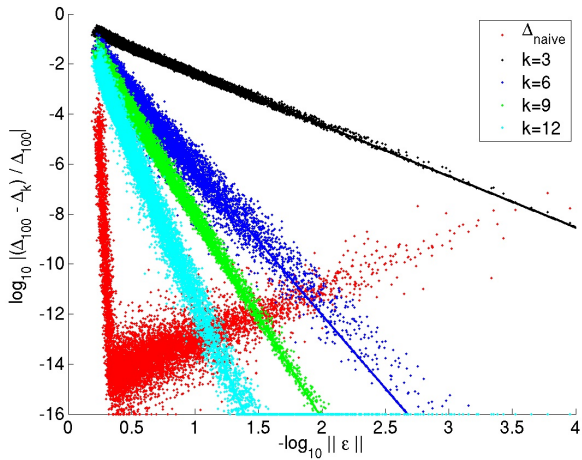
\caption{{Plot comparing the naive and approximate formulae, truncated at different
orders for calculating JSD as a function of the normalized L2-distance
($\| \beps \|$; see Section III) between pairs of randomly generated
probability distributions. Best fit slopes are: -2.05 ($k=3$), -5.89 ($k=6$), 
-8.14 ($k=9$), -11.91 ($k=12$) and -105.43 (comparing \rm{naive} with $k=100$).}} 
\label{compare}
\end{minipage}
\quad
\begin{minipage}[t]{0.485\linewidth}
\incgraph{negJSplot}
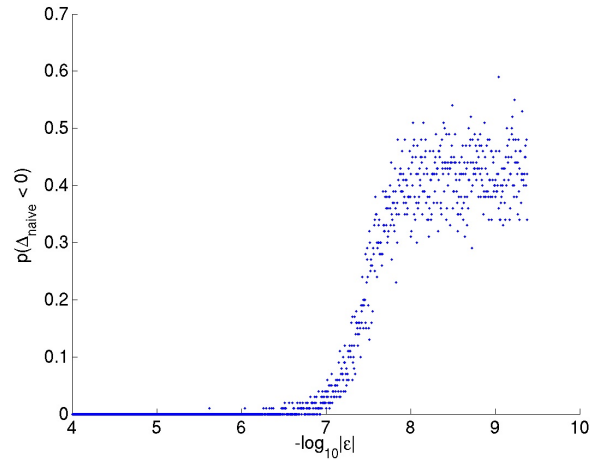
\caption{{Probability of obtaining (erroneous) negative values, when directly evaluating 
JSD using its exact expression, is plotted as a function of $\| \beps \|$. When implemented 
in \textsc{matlab}, we observe that the naive formula gives negative JSD when $\| \beps \|$
is merely of $O(10^{-6})$.}}
\label{negjs}
\end{minipage}
\efig

\appendix
\section*{Appendix}
Here we include the \textsc{matlab} code used in the figures for approximate evaluation of JSD using its series expansion. \\


\begin{verbatim}

function [JS,epsnorm] = JSapprx(pi1,p1,pi2,p2,order) 

% [JS,epsnorm]=JSapprx(pi1,p1,pi2,p2,order) calculates JS 
% divergence given two probability distributions and 
% their relative weights. JSapprx uses an approximation 
% to the JSD by expanding in powers of epsilon=(p1-p2)/(p1+p2)
% and truncating at an order input by the user. 
%
% This calculation is described in the technical report 
%       ``A non-negative expansion 
%       for small Jensen-Shannon Divergences'' 
% by Anil Raj and Chris H. Wiggins, October 2008 

% average of distributions 
pbar=(p1+p2)/2; 
% difference of distributions 
eta=(p1-p2)/2; 
% ratio of difference to average 
epsilon=eta./pbar; 
% difference in biases, where pi1+pi2=1 
alpha=pi1-pi2; 

% calculate JS by summing up to order `order' 
js=zeros(size(pbar)); 
% denominator computed by summing, as well 
denominator=0; 
for i=2:order 
  denominator=denominator+(i-1); 
  % numerical coefficient 
  c=(-1)^i*(1/denominator); 
  Bi=c*(alpha^(mod(i,2))-alpha^i); 
  js=js+Bi*(epsilon.^i); 
end 

% sum over `j': 
JS=pbar'*js/2; 

% convert from nats to bits: 
JS=JS/log(2); 

% norm of epsilon reported as output 
if nargout==2 
  epsnorm=sqrt(sum(epsilon.^2)/length(pbar)); 
end 

\end{verbatim}

\bibliographystyle{unsrt}
\bibliography{apprxJS}

\begin{thebibliography}{1}

\bibitem{lin:1991}
J~Lin.
\newblock Divergence measures based on the shannon entropy.
\newblock {\em IEEE Transactions on Information Theory}, 37(1):145--151, Jan
  1991.

\end{thebibliography}

\end{document}